# A Reinforcement Learning Approach to Health Aware Control Strategy

Mayank S Jha*, Philippe Weber, Didier Theilliol, Jean-Christophe Ponsart, Didier Maquin

*Abstract*—Health-aware control (HAC) has emerged as one of the domains where control synthesis is sought based upon the failure prognostics of system/component or the Remaining Useful Life (RUL) predictions of critical components. The fact that mathematical dynamic (transition) models of RUL are rarely available, makes it difficult for RUL information to be incorporated into the control paradigm. A novel framework for health aware control is presented in this paper where reinforcement learning based approach is used to learn an optimal control policy in face of component degradation by integrating global system transition data (generated by an analytical model that mimics the real system) and RUL predictions. The RUL predictions generated at each step, is tracked to a desired value of RUL. The latter is integrated within a cost function which is maximized to learn the optimal control. The proposed method is studied using simulation of a DC motor and shaft wear.

## I. INTRODUCTION

In the last decade, extensive research has been made in the domain of component-based Prognostics and Health Management (PHM). As almost all the industrial and mission critical systems operate in closed loop, new methods are being sought so that useful life of critical systems can be extended. In this context, health-aware control (HAC) has recently emerged as one of the domains where control synthesis is sought, based upon the state of health (SOH) and/or Remaining Useful Life (RUL) prognostics of critical components [1], [2]. Such control paradigms should ensure the stability, while satisfying the desired health-based constraints, plausibly by trading off present system performance at the expense of future health conditions [3].

Some of the existing methods are developed using reliability-based indicators such as [4]. For instance, [5], [6] proposed reconfiguration strategies in model predictive control (MPC) framework. On the other hand, RUL prognostics is obtained using Degradation models (DMs) that estimate their parameters using actual measurements and adapt the degradation dynamics based upon individual asset. This leads to efficient RUL predictions even in presence of large individual to individual variations in large fleet of assets. The degradation models can be constructed based upon several criteria such as energy loss [7], degradation physics [8] or statistical analysis of degradation data-base [9].

Reinforcement Learning (RL) based methods have seen a rapid surge in research mainly due to their ability to learn optimal policies (control laws) based upon interaction with the environment, in a model based as well as model-free setting. A general RL framework involves an actor agent (controller) that interacts with the environment (system or dynamic process) and produces a set of control actions. Then, based upon stimulus (response or reward) received from the environment, it modifies the control actions, eventually leading to learning of optimal control policy (control law). Recently, problems belonging to various scientific domains have been addressed in RL setting including feedback control and optimal adaptive control [10]. RL has been applied in various engineering domains including robotics [11] and recently for fault tolerant control[12]. RL based algorithms can be model based as well as model free. Typically, a Markov Decision Model (MDP) that represents the dynamics of the system is considered to generate the next state as well as reward, in order to learn the optimal control law. However, Q-function based RL algorithms are model-free in that the data obtained from process (as samples or trajectories) can be used to learn the inherent dynamics and the optimal control. This contrasts with the usual requirement of an analytical transition model (mathematical expression) in Control Community. Thus, RL based algorithms can be envisaged to provide answers for control synthesis problems in face of partial model knowledge, uncertainties or complete absence of system dynamics model.

For systems under degradation, the DMs of concerned components are usually available. For example, DMs of roller bearings are known to be exponential based upon offline processing of degradation data. However, exact knowledge of the global system (usually complex) is usually not available. Moreover, in order to generate health/RUL predictions, a suitable prediction model is required. Such mathematical dynamic (transition) models of RUL are rarely available which makes it difficult for RUL information to be incorporated into the control paradigm. This calls for novel methods that envisage integration of databases (data-driven methods) with model-based knowledge for control learning.

In this paper, a novel method is proposed wherein model-based knowledge of system and component degradation, is integrated with RUL prediction data within Reinforcement Learning (RL) framework in order to obtain optimal control policy. To that end, the RUL predictions generated at each step is tracked to a desired value of RUL and the latter is integrated within a cost function (reward), which is maximized to learn the optimal control. Clearly, the scientific interest remains in developing new RL based methods for rendering intelligence to HAC. This is a novel attempt towards integrating model-based approach with model-free database (RUL predictions) to learn the optimal control for HAC. Moreover, it is novel in that this methodology renders intelligence (learning ability) for prognostics-based control. This section is followed by description of problem formulation. Subsequently, a framework is proposed. Finally, simulation results are presented followed by conclusions

Authors are with Centre de Recherche en Automatique de Nancy (CRAN), UMR 7039, CNRS, University of Lorraine, 54506 Vandoeuvre Cedex, France (*corresponding author, e-mail: mayank-shekhar.jha@univ-lorraine.fr).

## II. PROBLEM FORMULATION

The optimal control learning problem can be formalized by using Markov Decision Process (MDP) model [13] of the dynamical system. The deterministic setting is considered in this paper with $X$ being state space of the process, $U$ being the action space, $f_s$ being the state transition function. Consider a general nonlinear system affine in control in discrete time $k$:

$$x_{k+1} = f_s(x_k) + g(x_k)u_k \quad (1)$$

where $f_s$ describes system evolution (plausibly non-linear) and $g$ describes the input evolution, with system variables $x_k \in X$, control input $u_k \in U$. Then, the control policy $h(\cdot): X \to U$ is defined as function from state space to control space as:

$$u_k = h(x_k) \quad (2)$$

Depending on the control action $u_k$, the state transits to $x_{k+1}$ and receives a scalar reward $r_{k+1}$ generated by reward function $\rho: X \times U \to \mathbb{R}$ as:

$$r_{k+1} = \rho(x_k, u_k) \quad (3)$$

Equations (1) and (3) along with control policy, define the MDP. It is highlighted that given $x_k$, $u_k$, $f$ and $\rho$, the next state $x_{k+1}$ and reward $r_{k+1}$ can be determined using the Markov property as the successive state $x_{k+1}$ only depends upon $x_k$ and $u_k$. MDP model formalization enables us to take advantage of RL based methods for optimal control generation.

In Fig.1, the global architecture is shown along with the module containing system dynamics and degradation model module.

*Assumption 1:* System variables including those sensitive to degradation are considered observable (estimated).

*Assumption 2:* The system is considered stabilizable on some set $\Omega_x \in X$ which implies that there exists a admissible control policy $h(x_k)$, such that the closed loop system $x_{k+1} = f_s(x_k) + g(x_k)h(x_k)$ is asymptotically stable on $\Omega$.

### A. RUL Prediction Module

The degradation dynamics is not embedded into the system dynamic model(1), but processed in a separate computational block which considers the degradation of critical component/subsystem (chosen a priori). Clearly, this formulation is context/system dependent. It is motivated by the fact that DMs are usually statistically derived using offline controlled environment tests (non-destructive testing) and/or obtained from identification/processing of features in temporal/frequential feature-space. The degradation is considered for a subsystem/component as:

$$d_{k+1} = f_d(d_k, m, x_k) \quad (4)$$

where $d \in D$ represents any degradation index that evolves according to function (plausibly non-linear) $f_D$, depends on a set of known time invariant parameters (plausibly estimated beforehand) $m \in M$ and a set of system variables $x_k$. The initial value of degradation index $d_0$ is considered known (plausibly estimated beforehand).

*Assumption 3:* Given that system variables sensitive to DM (4) are stabilizable, the DM is considered stabilizable over some set $\Omega_d \in D$.

### B. L-Step Ahead Prediction For RUL

Given $D_{fail}$ as the failure value of degradation index, the RUL at any discrete time instant $k$ can be obtained using *l-step* ahead prediction. This is done by projecting the DM (4) into future over an infinite horizon [14], [15], till $D_{fail}$ is reached. Such a projection involves *l-step* iteration of (1) and (4) into future. In this work, we generate *l-step* ahead prediction at instant $k$, by assuming future control inputs $u_{k+i}, i \in (0,..,l)$ same as $u_k$. When $d_{k+l} = D_{fail}$, the simulation is stopped, and number of steps taken to reach the future equals the RUL (in discrete time) at $k$.

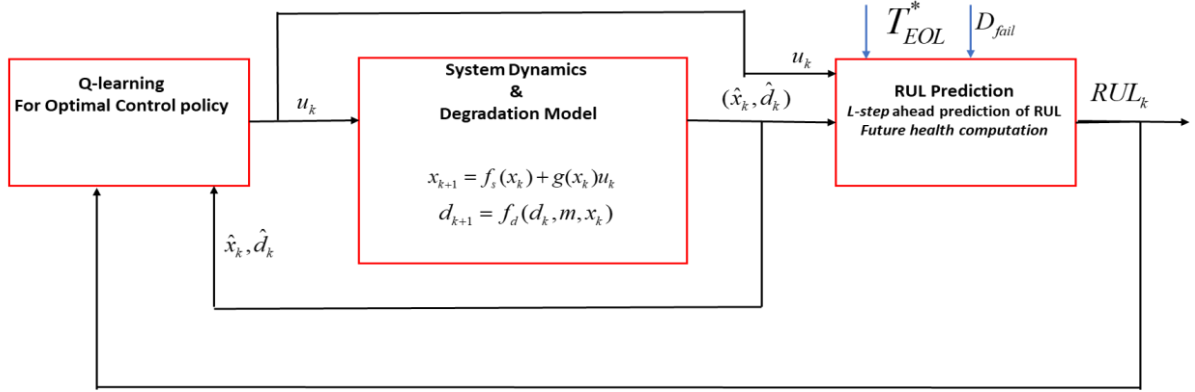

Figure 1 Health aware control framework

*Assumption 4:* $D_{fail}$ is considered within the stabilizable domain $\Omega_d$ or $D_{fail} \in \Omega_d$. Assumption 4 implies that there exists a stabilizing control at any instant $k$ that leads to a finite RUL value at $k$. In principle, definition of *RUL* is meaningful till the true end of life (EOL) is reached.

### C. Desired-RUL Trajectory Generation

For mission-critical systems with component degradation, it is usual to have a *desired end of life* $T_{EOL}^*$ set *a priori*. This value is usually mission dependent and user based. Consider the *end of life* of the considered system as $T_{EOL}$. Clearly, the RUL at $k$ is the amount of time left to reach $T_{EOL}$, or:

$$RUL_k = T_{EOL} - k \times Ts \tag{5}$$

where *Ts* is the sampling time. Thus, the *desired* RUL at time $k$, $RUL_k^*$ can be expressed as:

$$RUL_k^* = T_{EOL}^* - k \times Ts \tag{6}$$

Thus, at any step $k$, $RUL_k$ can be calculated through *l*-step ahead prediction and $RUL_k^*$ can be obtained from (6). Availability of $RUL_k^*$ motivates formulation of the problem as an optimal RUL tracking problem.

**Remark:** The knowledge of $T_{EOL}^*$ is user based/mission-imposed, which can in turn be used to generate a desired trajectory from (6). We formulate the problem as a Q-learning problem to learn optimal control that assures efficient RUL tracking.

### D. Optimal RUL tracking

Given $x_k, RUL_k, u_k$, a reward $r_{k+1}$ is generated as:

$$r_{k+1} = -\frac{1}{2}\left(x_k^T S x_k + u_k^T R u_k + (RUL_k^* - RUL_k)^T P(RUL_k^* - RUL_k)\right) \tag{7}$$
$$= \rho(x_k, RUL_k, u_k)$$

where S, R and P are invariant positive semi-definite matrices of appropriate sizes. The reward function $\rho(\cdot)$ is quadratic in nature, inspired from Linear Quadratic regulator type cost function such that argument of $\rho(\cdot)$ are minimized as reward $r_{k+1}$ is maximized over stages. Thus, we seek to minimize the difference $(RUL_k^* - RUL_k)$ while assuring minimal energy consumption as well as desired system performance.

It is noted that system (1), control policy (2) and reward (7), constitute the MDP. Cumulative reward or *return* $R^h(x_k, RUL_k, u_k)$ from system state $x_k, RUL_k$, following the control policy $u_k = h(x_k)$, is:

$$R^h(x_k, RUL_k, u_k) = \sum_{i=k}^{\infty} \gamma^{i-k} r_{i+1} = \sum_{i=k}^{\infty} \gamma^{i-k} \rho(x_i, RUL_i, h(x_i)) \tag{8}$$

where $\gamma$ such that $0 < \gamma < 1$ is the discount factor over the infinite horizon. The return is maximized over an infinite horizon which allows us to obtain stationary policies. The cost or value associated with a control policy is assessed by two types of 'value' functions: state value functions (V-functions) and state-action value functions (Q-functions) [16]. Q-function is defined with respect to both state $x_k$ and action $u_k$, as the function that generates return obtained by applying $u_k$ at state $x_k$ and following policy $h(x_k)$ thereafter. The Q-function then becomes:

$$Q^h(x_k, RUL_k, u_k) = \rho(x_k, RUL_k, u_k) \tag{9}$$
$$+ \gamma R^h(x_{k+1}, RUL_{k+1}, h(x_{k+1}))$$

where $RUL_{k+1}$ is produced by *l-step ahead prediction* by following the same policy $h(\cdot)$ at next system state $x_{k+1}$. Bellman's equation for Q-function is:

$$Q^h(x_k, RUL_k, u_k) = \rho(x_k, RUL_k, u_k) \tag{10}$$
$$+ \gamma Q^h(x_{k+1}, RUL_{k+1}, h(x_{k+1}))$$

Further, optimal Q-function $Q^*$ is the one that maximizes Q-function for any policy, or, $Q^*(x, RUL, u) = \max_{u \in U} Q^h(x, RUL, u)$. Bellman optimal equation for Q-functions can be obtained as:

$$Q^*(x, u) = \rho(x, u) + \gamma \max_{u'} Q^*(f_s(x, u), u') \tag{11}$$

Clearly, the optimal policy $h^*$ is the one that selects at each state, an action with largest optimal Q-value, or:

$$h^*(x) = \arg\max_{u \in U} Q^*(x, RUL, u) \tag{12}$$

*Optimality:* It has been shown that Bellman equations are fixed point equations, which translates to the fact that starting from an arbitrary $Q_0$, a unique fixed point is found upon successive iterations in asymptotic sense as (11) is a contraction mapping with factor $\gamma < 1$ in infinity norm; it converges to optimal $Q^*$ asymptotically as $j \to \infty$ or $\|Q_{j+1} - Q^*\|_\infty \leq \gamma \|Q_j - Q^*\|_\infty$

*Model-free Q-learning:* In order to avoid the need of transition function $f_s$, Q-iteration is performed in model free sense using only the observed state transitions and rewards, i.e., data tuples $(x_k, RUL_k, u_k, x_{k+1}, RUL_{k+1}, r_{k+1})$. The problem is cast in RL framework under class of model free value iteration algorithms. We use *Q-learning* algorithm which starts from an arbitrary initial Q-function and updates it as:

$$Q_{k+1}(x_k, RUL_k, u_k) = Q_k(x_k, RUL_k, u_k)$$
$$+ \alpha_k \left[ \underbrace{r_{k+1} + \gamma \max_{u'} Q_k(x_{k+1}, RUL_{k+1}, u')}_{\text{updated estimate}} - \underbrace{Q_k(x_k, u_k)}_{\text{current estimate}} \right] \tag{13}$$
$$\underbrace{\phantom{xx}}_{\text{temporal difference}}$$

where $\alpha_k \in [0,1]$ is the learning rate and the term inside bracket represents the *temporal difference*, i.e., difference between $r_{k+1} + \gamma \max_{u'} Q_k(x_{k+1}, RUL_{k+1}, u')$: the updated estimate of the optimal Q-value of $Q_k(x_k, RUL_k, u_k)$, and current estimate of $Q_k(x_k, RUL_k, u_k)$. In order to ensure that whole state space is visited (exploration) by control actions in asymptotic sense, we employ a $\varepsilon$-greedy exploration strategy wherein control action is selected according to:

$$u_k = \begin{cases} u \in \arg\max_{u \in U} Q(x, RUL, u) \text{ with probability } 1-\varepsilon_k \text{ (exploitation)} \\ u \in \text{random}(U) \quad \text{with probability } \varepsilon_k \quad \text{(exploration)} \end{cases} \quad (14)$$

where $\varepsilon \in (0,1)$ is the exploration probability at step $k$. In this paper, the epsilon-decay routine is considered such that with each iteration as:

$$\varepsilon = \frac{1}{number\ of\ episodic\ play\ \text{(iteration)}}.$$

### III. SIMULATIONS

We consider a second order DC (direct current) motor model as our system, obtained by discretizing a continuous model with sampling time $Ts=0.01s$.

$$\begin{bmatrix} i_{k+1} \\ \omega_{k+1} \end{bmatrix} = \begin{bmatrix} 0.9 & -0.001 \\ 0.001 & 0.99 \end{bmatrix} \begin{bmatrix} i_k \\ \omega_k \end{bmatrix} + \begin{bmatrix} 0.01 \\ 0 \end{bmatrix} u_k \quad (15)$$

A simplified model of wear is considered to model localized wearing of shaft due to load and angular rotation. Using Archard equation [17], the wear rate $Hw$ (m³s⁻¹) is considered as function of shaft speed as first order differential of wear $\frac{d(Hw)}{dt} = C_w \cdot \omega(t)$. It is considered in discrete time $k$ as:

$$Hw_{k+1} = Hw_k + Ts \cdot C_w \cdot \omega_k \quad (16)$$

where the wear coefficient $C_w=0.1$ m²/N, is assumed known. Thus, a simplified linear degradation model with respect to wear mechanism is considered with a known initial value. Using saturation, the shaft speed is bounded within $\omega \in [0, 10^{-1}]$ rad s⁻¹, output current as $i(t) \in [0, 0.1]$ A and input voltage is considered bounded as $u(t) \in [0, 10]$ V. Clearly, system configurations are chosen in a way such that all significant variables are within similar scale-ranges for the ease of analysis and simulation.

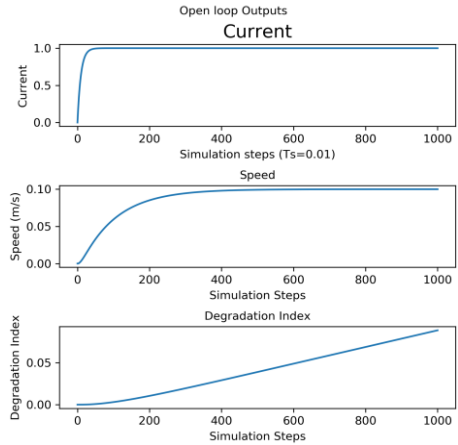

Figure 2 Open loop characteristics

Figure 2 shows the open loop characteristics obtained with input of 10V from **one *episodic*** simulation run, which consists of 1000 simulation steps (10s).

Failure value of wear rate is set to $Hw_{fail} = 0.02$ (m³/s), which is attained at around 3.4s under open loop and full loading conditions, i.e. $T_{EOL} = 3.4s$. The objective is to reach a desired $T^*_{EOL}=9s$ by learning a suitable control law. As sampling time $Ts=0.01s$, $T^*_{EOL}=9s$ corresponds to 900 simulation steps in time. Thus, the objective becomes learning suitable control law that allows $Hw$ reach not more than $0.02$(m³/s) at the end of 900 simulation steps, thereby producing 0 RUL at the end of 900 iterations. At start, we have the desired $RUL^*(k=0) = T^*_{EOL}$ and further on, $RUL^*_k = T^*_{EOL} - k \times Ts$. The rewards are proposed as:

$$r_{k+1} = \qquad (17)$$
$$-\frac{1}{2}\left( \begin{bmatrix} i_k \\ \omega_k \end{bmatrix}^T \begin{bmatrix} 0.1 & 0 \\ 0 & 0.1 \end{bmatrix} \begin{bmatrix} i_k \\ \omega_k \end{bmatrix} + (10)u_k^2 + (100)(RUL^*_k - RUL_k)^2 \right)$$

Clearly, the reward structure intends to incentivize (ten times more) the minimization of term $(RUL^*_k - RUL_k)$ over the others. For example, a greater negative reward is given when desired RUL is NOT reached at any step. Thus, the agent (via Q-learning) learns to minimize this difference since maximum of the return (cumulative reward) is considered during Q-learning. For Q-learning, a naïve tabular approach is adopted in this paper such that state-action pair associated with each Q-value is stored in a table. To this end, the state space of Q-function which comprises of state variables of the system and RUL, is partitioned into equal number of bins. Each of the Q-function variable, $i(t) \in [0,1]$ A, $\omega(t) \in [0, 10^{-1}]$ rad s⁻¹ and $RUL(t) \in [0, 200]$ s is discretized into 50 bins, leading to $50^3=125000$ discretized spaces along the same axis. Then, the input space $u(t) \in [0, 10]$ volts is divided into 20 equal bins along second axis leading to a 2-D tabular space of size 125000 X 20. During each iteration of episodic play (simulation of 1000 steps), The Q-value associated with each state-action pair is stored and updated in tabular form. The discount factor $\gamma = 0.95$ is considered and $\varepsilon = 0.1$ is taken for the $\varepsilon$-greedy approach. The epsilon values decays with each episodic play (after one simulation run). The Q-values were initialized randomly and for each $k$ within an episode (of 10 seconds), the generated data tuples $(x_k, RUL_k, u_k, x_{k+1}, RUL_{k+1}, r_{k+1})$ were used to produce a reward and update the Q-function for all the states visited during that episodic run.

It should be noted that for an episode, at each time step, DC motor model plays the role of a real system by providing the state data (states considered observable), and DM generates the RUL prediction, followed by update of the Q-values (state-action pairs) stored in a table.

A total of 3000 episodic runs (of 1000 simulation time steps each) were performed with initial values of variables remaining same for each episode (exploring-starts approach was not adopted). Convergence of rewards and Q-values for all states, was detected after approx. 1500 episodic runs as seen in Fig. 3. Fig. 4 shows the system response, degradation levels, the RUL under the control law learnt as well as the learnt control law itself (also see Fig. 6).

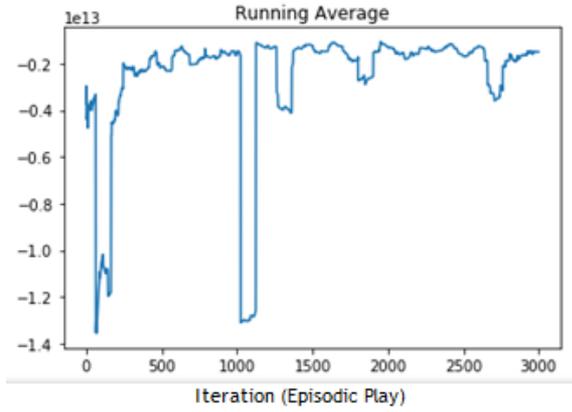

Figure 3 Running Average of Rewards after each 10 iterations (10 episodic plays)

Clearly, the degradation levels is restricted to $Hw_{fail}$ value at the end of 10s or 1000 simulation steps. Fig. 5 shows the zoomed profile of RUL. As shown, the control law tracks the desired RUL in such a way that approx. zero value is reached at the end of 9s of system functioning (900 simulation steps) leading to the approx. desired RUL.

The simulations were performed on Intel Xeon processor with 32GB of RAM and 3.0GHz of clock frequency. Also, the simulations were programed in Python 3.6.

## V. CONCLUSION

A novel framework for health aware control is presented in this paper where reinforcement learning is used to learn an optimal control policy in face of component degradation by integrating global system transition data (generated by an analytical model that mimics the real system) and RUL prediction data, in reinforcement learning framework. This leads to learning of suitable control that manages the speed of degradation in a way such that desired RUL is reached.

Clearly, the control is learnt in an offline way. Around 1500 episodic plays (dynamic simulations of 100 steps) were needed in order to converge towards optimal control. Once an appropriate control is learnt it can be applied in online manner.

Moreover, in absence of accurate dynamics of system, this can be achieved by storing system transition data and using *experience replay* in order to learn the control in offline manner. This paper assumes availability of observable states.

In further works, estimation of hidden states will be investigated. Moreover, this method needs to be developed to handle highly non-linear degradation process which are often stochastic in nature. A naïve tabular approach was adopted in this paper which supports the preliminary results.

The quality of control can be made better using efficient function approximators for control learning. In future works, author intend to develop more efficient function approximation methods to handle large state spaces and complex system dynamics.

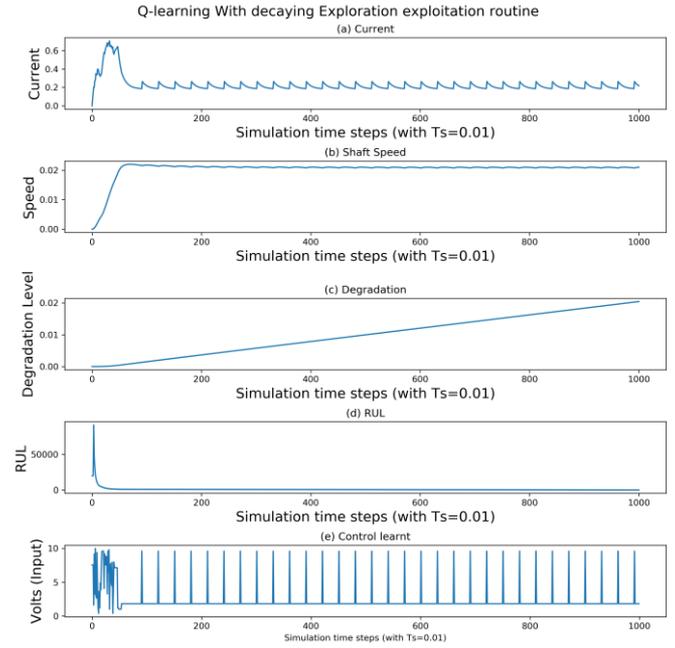

Figure 4 (a) & (b) System response, (c) degradation state profile, (d) RUL and (e) system input command learnt using the proposed Q-learning procedure

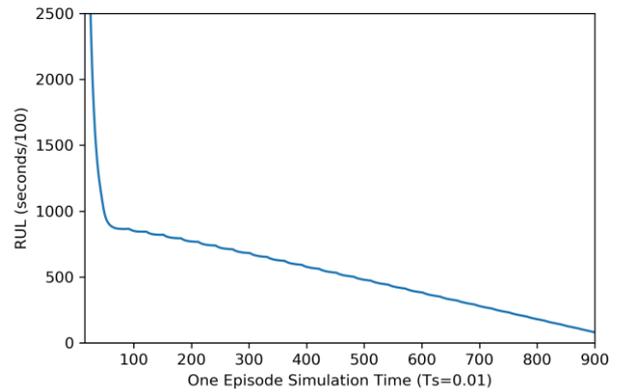

Figure 5 RUL under command learnt (zoomed version)

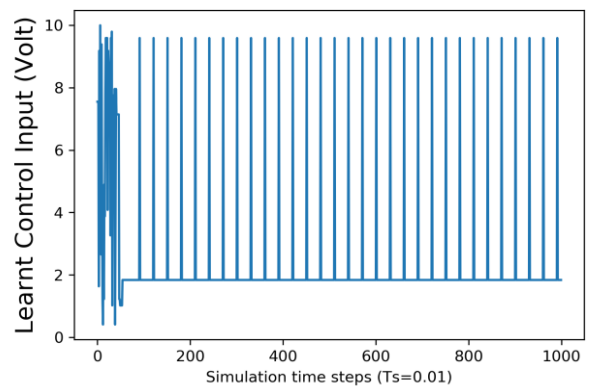

Figure 6 Control input (Volts) learnt